\documentclass[runningheads]{llncs}
\usepackage{graphicx}
\usepackage{comment}
\usepackage{amsmath,amssymb} 
\usepackage{color}
\usepackage{array}
\usepackage{tabu}
\usepackage{esvect}

\newcommand{\vpara}[1]{\vspace{0.1in}\noindent\textbf{#1 }} 
\newcommand{\norm}[1]{\left\lVert#1\right\rVert}

\begin{document}
\pagestyle{headings}
\mainmatter
\def\ECCVSubNumber{1220}  

\title{Coupling Explicit and Implicit Surface Representations for Generative 3D Modeling}

\author{Omid Poursaeed$^{1,2}$ \and
Matthew Fisher$^{3}$ \and
Noam Aigerman$^{3}$ \and
Vladimir G. Kim$^{3}$ 
     \newline  
\institute{$^1${Cornell University}\qquad
    $^2${Cornell Tech}\qquad
    $^3${Adobe Research}
    }}

\titlerunning{Coupling Explicit and Implicit Surface Representations}
%

%
\authorrunning{O. Poursaeed, M. Fisher, N. Aigerman and V. Kim}
%
\maketitle

\begin{abstract}
We propose a novel neural architecture for representing 3D surfaces, which harnesses two complementary shape representations: (i) an explicit representation via an atlas, i.e., embeddings of 2D domains into 3D; (ii) an implicit-function representation, i.e., a scalar function over the 3D volume, with its levels denoting surfaces. We make these two representations synergistic by introducing novel consistency losses that ensure that the surface created from the atlas aligns with the level-set of the implicit function. Our hybrid architecture outputs results which are superior to the output of the two equivalent single-representation networks, yielding smoother explicit surfaces with more accurate normals, and a more accurate implicit occupancy function. Additionally, our surface reconstruction step can directly leverage the explicit atlas-based representation. This process is computationally efficient, and can be directly used by differentiable rasterizers, enabling training our hybrid representation with image-based losses.
 
\end{abstract}

\section{Introduction}

Many applications rely on a neural network to generate a 3D geometry~\cite{Han2019SVRsurvey,dai2017complete}, where early approaches used point clouds~\cite{achlioptas2018latent_pc}, uniform voxel grids~\cite{liu2018voxelgan}, or template mesh deformations~\cite{benhamu2018multichart} to parameterize the outputs. The main disadvantage of these representations is that they rely on a pre-selected discretization of the output, limiting network's ability to focus its capacity on high-entropy regions. Several recent geometry learning techniques address this limitation by representing 3D shapes as \emph{continuous} mappings over vector spaces. Neural networks learn over a manifold of these mappings, creating a mathematically elegant and visually compelling generative models. Two prominent alternatives have been proposed recently.




The explicit surface representation defines the surface as an atlas -- a collection of \emph{charts}, which are maps from 2D to 3D, $\{ f_i : \Omega_i\subset \mathbb{R}^2 \rightarrow \mathbb{R}^3 \}$, with each chart mapping a 2D patch $\Omega_i$ into a part of the 3D surface. the surface $S$ is then defined as the union of all 3D patches, $S = \cup_i f_i\left(\Omega_i\right)$.
In the context of neural networks, this representation has been explored in a line of works considering atlas-based architectures~\cite{groueix2018atlasnet,yang2018foldingnet} which exactly represent surfaces by having the network predict the charts $\{f_i^{\bf{x}}\}$, where the network also takes latent code, $\bf{x} \in \mathcal{X}$, as input, to describe the target shape. These predicted charts can then be queried at arbitrary 2D points, enabling approximating the resulting surface with, e.g., a polygonal mesh, by densely sampling the 2D domain with the vertices of a mesh, and then mapping the resulting mesh to 3D via $f_i^{\bf{x}}$. This reconstruction step is suitable for an end-to-end learning pipeline where the loss is computed over the resulting surface. It can also be used as an input to a differentiable rasterization layer in case image-based losses are desired. On the other hand, the disadvantage of atlas-based methods is that the resulting surfaces tend to have visual artifacts due to inconsistencies at patch boundaries. 

The implicit surface representation defines a volumetric function $g:\mathbb{R}\to\mathbb{R}^3$. This function is called an implicit function, with the surface $S$  defined as its zero level set, $S = \{ p \in \mathbb{R}^3 | g\left(p\right) = 0 \}$.
Many works train networks to predict implicit functions, either as signed distance fields~\cite{park2019deepsdf,chen2019implicitfields}, or simply occupancy values \cite{mescheder2019occupancy}. They also typically use shape descriptor, ${\bf \hat{x}}\in \hat{\mathcal{X}}$, as additional input to express different shapes: $g^{\bf \hat{x}}$. 
These methods tend to produce visually appealing results since they are smooth with respect to the 3D volume. They suffer from two main disadvantages; first, they do not immediately produce a surface, making them less suitable for end-to-end pipeline with surface-based or image-based losses; second, as observed in~\cite{park2019deepsdf,chen2019implicitfields,mescheder2019occupancy}, their final output tends to produce a higher surface-to-surface distance to ground truth than atlas-based methods. 

In this paper we propose to use both representations in a hybrid manner, with our network predicting both an explicit atlas $\{f_i\}$ and an implicit function $g$. For the two branches of the two representations we use the AtlasNet~\cite{groueix2018atlasnet} and OccupancyNet~\cite{mescheder2019occupancy} architectures. We use the same losses used to train these two networks (chamfer distance and occupancy, respectively) while adding novel consistency losses that couple the two representations during joint training to ensure that the atlas embedding aligns with the implicit level-set.
We show the two representations reinforce one another: OccupancyNet learns to shift its level-set to align it better with the ground truth surface, and AtlasNet learns to align the embedded points and their normals to the level-set. This results in smoother normals that are more consistent with the ground truth for the atlas representation, while also maintaining lower chamfer distance in the implicit representation. Our framework enables a straightforward extraction of the surface from the explicit representation, as opposed to the more intricate marching-cube-like techniques required to extract a surface from the implicit function. This enables us to add image-based losses on the output of a differentiable rasterizer. Even though these losses are only measured over AtlasNet output, we observe that they further improve the results for \emph{both} representations, since the improvements propagate to OccupancyNet via consistency losses. Another advantage of reconstructing surfaces from the explicit representation is that it is an order of magnitude faster than running marching cubes on the implicit representation.
We demonstrate the advantage of our joint representation by using it to train 3D-shape autoencoders and reconstruct a surface from a single image. The resulting implicit and explicit surfaces are consistent with each other and quantitatively and qualitatively superior to either of the branches trained in isolation.

\section{Related Work}

We review existing representations for shape generation that are used within neural network architectures. While target application and architecture details might vary, in many cases an alternative representation can be seamlessly integrated into an existing architecture by modifying the layers of the network that are responsible for generating the output. 

Generative networks designed for images operate over regular 2D grids and can directly extend to 3D voxel occupancy grids~\cite{liu2018voxelgan,DBLP:journals/corr/GirdharFRG16,DBLP:journals/corr/BrockLRW16,dai2017complete}. These models tend to be coarse and blobby, since the size of the output scales cubically with respect to the desired resolution. Hierarchical models~\cite{hane2017hierarchical,riegler2017octnet} alleviate this problem, but they still tend to be relatively heavy in the number of parameters due to multiple levels of resolution. A natural remedy is to only focus on surface points, hence point-based techniques were proposed to output a tensor with a fixed number of 3D coordinates~\cite{achlioptas2018latent_pc,Su2017PointGen}. Very dense point clouds are required to approximate high curvature regions and fine-grained geometric details, and thus, point-based architectures typically generate coarse shapes. While polygonal meshes allow non-even tessellation, learning over this domain even with modest number of vertices remains a challenge~\cite{dai2019scan2mesh}. One can predict vertex positions of a template~\cite{liu2018meshVAE,poursaeed2020neural}, but this can only apply to analysis of very homogeneous datasets. Similarly to volumetric cases, one can adaptively refine the mesh~\cite{wang2018pixel2mesh,2019arXiv191203681W} using graph unpooling layers to add more mesh elements {or iteratively refine it via graph convolutional networks~\cite{wen2019pixel2mesh++}. 
This refinement can be conditioned on images~\cite{wang2018pixel2mesh,wen2019pixel2mesh++} or 3D volumes~\cite{2019arXiv191203681W}.} 
The main limitation of these techniques is that they discretize the domain in advance and allocate same network capacity to each discrete element. 
Even hierarchical methods only provide opportunity to save time by not exploring finer elements in feature-less regions. In contrast, continuous, functional representations enable the network to learn the discretization of the output domain. 

The explicit continuous representations view 3D shapes as 2D charts embedded in 3D~\cite{groueix2018atlasnet,yang2018foldingnet}. These atlas-based techniques tend to have visual artifacts related to non-smooth normals and patch misalignments. 
For homogeneous shape collections, such as human bodies, this can be remedied by replacing 2D charts with a custom template (e.g., a human in a T-pose) and enforce strong regularization priors (e.g., isometry)~\cite{groueix20183d}, however, the choice of such a template and priors limits expressiveness and applicability of the method to non-homogeneous collections with diverse geometry and topology of shapes. 

Another alternative is to use a neural network to model a space probing function that predicts occupancy~\cite{mescheder2019occupancy} or clamped signed distance field~\cite{park2019deepsdf,chen2019implicitfields} for each point in a 3D volume. Unfortunately, these techniques cannot be trained with surface-based losses and thus tend to perform worse with respect to surface-to-surface error metrics. 

Implicit representations also require marching cubes algorithm~\cite{Lorensen87} to reconstruct the surface. Note that unlike explicit representation, where every sample lies on the surface, marching cubes requires sampling off-surface points in the volume to extract the level set. We found that this leads to a surface reconstruction algorithm that is about an order of magnitude slower than an explicit technique. 
This additional reconstruction step, also makes it impossible to plugin the output of the implicit representation into a differentiable rasterizer (e.g., ~\cite{Loper:ECCV:2014,kato2018renderer,liu2019softras}. We observe that using differentiable rasterizer to enforce additional image-based losses can improve the quality of results. Moreover, adding these losses just for the explicit output, still propagates the improvements to the implicit representation via the consistency losses. 

In theory, one could use differentiable version of marching cubes~\cite{Liao2018CVPR} for reconstructing a surface from an implicit representation, however, this has not been used by prior techniques due to cubic memory requirements of this step (essentially, it would limit the implicit formulation to $32^3$ grids as argued in prior work~\cite{mescheder2019occupancy}). Several recent techniques use ray-casting to sample implicit functions for image-based losses. Since it is computationally intractable to densely sample the volume, these methods either interpolate a sparse set of samples~\cite{liu2019renderimplicits} or use LSTM to learn the ray marching algorithm~\cite{sitzmann2019srns}. Both solutions are more computationally involved than simply projecting a surface point using differentiable rasterizer, as enabled by our technique.

%


\section{Approach}
\begin{figure*}[t]
\begin{center}
   \includegraphics[width=0.95\linewidth]{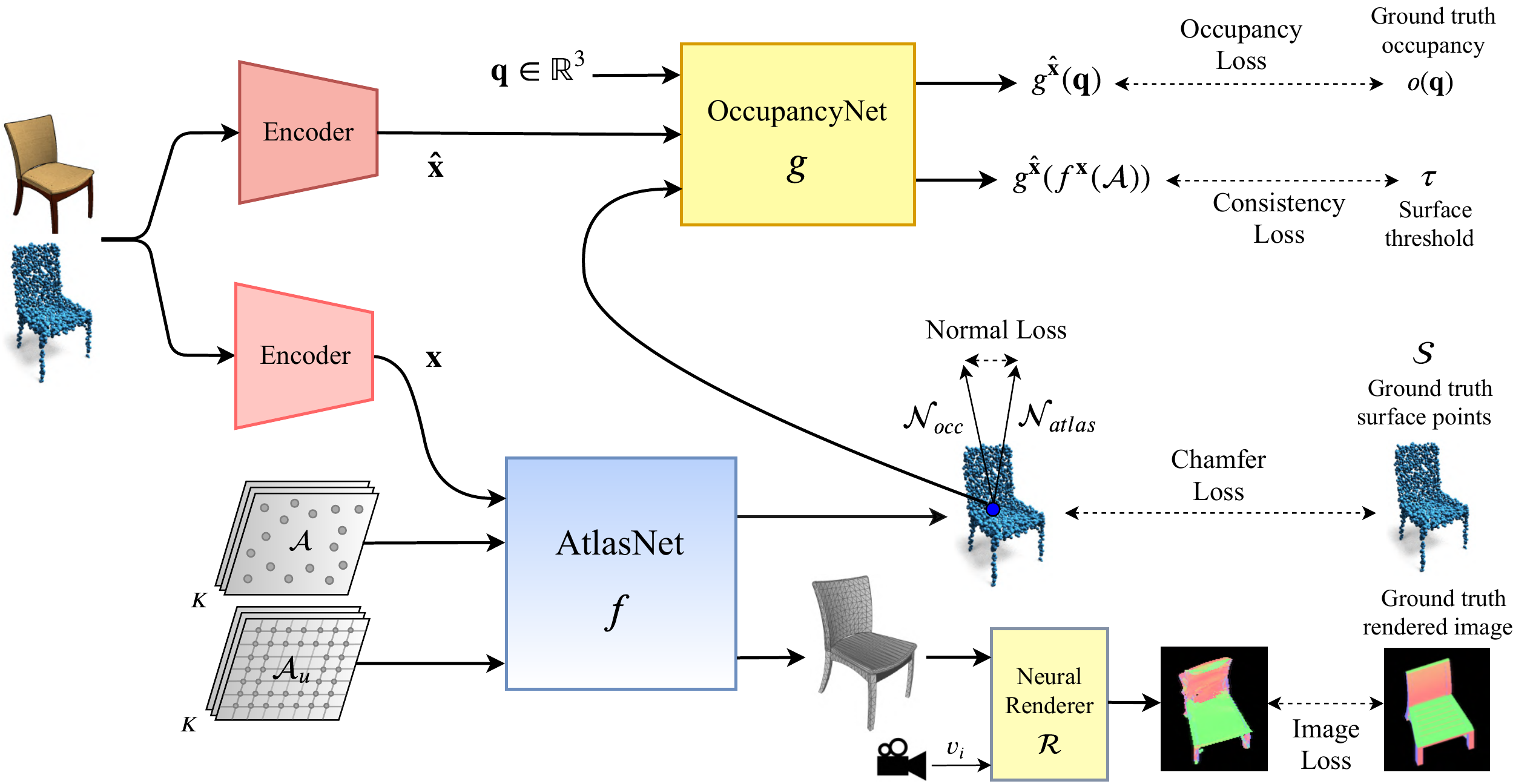}
\end{center}
   \caption{Model architecture. AtlasNet and OccupancyNet branches of the hybrid model are trained with Chamfer and occupancy losses as well as consistency losses for aligning surfaces and normals. A novel image loss is introduced to further improve generation quality. }\label{fig:architecture}
\end{figure*}

We now detail the architecture of the proposed network, as well as the losses used within the training to enforce consistency across the two representations.

\subsection{Architecture} Our network simultaneously outputs two surface representations. These two representations are generated from two branches of the network, where each branch uses a state-of-the-art architecture for the target representation. 

For the explicit branch, we use AtlasNet~\cite{groueix2018atlasnet}. AtlasNet represents $K$ charts with neural functions $\{f_i^{\bf{x}}\}_{i=1}^K$, where each function takes a shape descriptor vector, $\bf{x} \in \mathcal{X}$, and a query point in the unit square, ${\bf p}\in\left[0,1\right]^2$, and outputs a point in 3D, i.e., $f_i^{\bf{x}}:\left[0,1\right]^2\to\mathbb{R}^3$. We also denote the set of 3D points achieved by mapping all 2D points in $\mathcal{A}\subset \left[0, 1\right]^2$ as $f^{\bf x}\left(\mathcal{A}\right)$.

For the implicit branch, we use OccupancyNet~\cite{mescheder2019occupancy}, learning a neural function $g^{\bf \hat{x}}:\mathbb{R}^3\to[0, 1]$,  which takes a query point $q\in\mathbb{R}^3$ and a shape descriptor vector ${\bf \hat{x}}\in \hat{\mathcal{X}}$ and outputs the occupancy value. The point $q$ is considered occupied (i.e., inside the shape) if $g^{\bf \hat{x}} \geq \tau$, where we set $\tau = 0.2$ following the choice of OccupancyNet.

\subsection{Loss Functions} 
Our approach centers around losses that ensure geometric consistency between the output of the OccupancyNet and AtlasNet modules. We employ these consistency loses along with each branch's original fitting loss (Chamfer and occupancy loss) that was used to train the network in its original paper. Furthermore, we take advantage of AtlasNet's output lending itself to differentiable rendering in order to incorporate a rendering-based loss. These losses are summarized in Figure~\ref{fig:architecture} and detailed below.

\vpara{Consistency losses.}
First, to favor consistency between the explicit and implicit representations, we observe that the surface generated by AtlasNet should align with the $\tau$-level set of OccupancyNet:
\begin{equation}\label{eq:constraint}
    g^{\bf \hat{x}}\left(f_i^{\bf x}({\bf p})\right) = \tau,
\end{equation}
for all charts $f_i^{\bf x}$ and at every point ${\bf p}\in[0,1]^2$. Throughout this subsection we assume that ${\bf x}$ and ${\bf \hat{x}}$ are describing the same shape. 

This observation motivates the following surface consistency loss:
\begin{equation}\label{eq:consistency}
    \mathcal{L}_\text{consistency} = \sum_{{\bf p}\in \mathcal{A}}\mathcal{H}\big(g^{\bf \hat{x}}(f_i^{\bf x}({\bf p})), {\tau}\big).
\end{equation}
where $\mathcal{H}(\cdot, \cdot)$ is the cross entropy function, and $\mathcal{A}$ is the set of sample points in $[0,1]^2$. {More specifically, for each point $p_i$ sampled on a 2D patch and its mapped version $g(f(p_i))$, we measure the binary cross-entropy $\tau\log(g(f(p_i))) + (1-\tau)\log(1-g(f(p_i)))$ and sum the losses for all points in the batch. Therefore, the minimum occurs at $g(f(p_i))=\tau$ for all $i$.} 
{Note that the current loss function only penalizes surface points that are not on the level set of the implicit function, but does not penalize if the OccupancyNet has a zero level set far away from the AtlasNet surface. 
Such a loss can be added via differentiable layers that either extract the level set of the OccupancyNet or convert AtlasNet surface to an implicit function. Finding a computationally efficient way of incorporating this loss function is a good venue for future work.}

Second, we observe that the gradient of the implicit representation should align with the surface normal of the explicit representation. The normals for both representations are differentiable and their analytic expressions can be defined in terms of gradients of the network. For AtlasNet, we compute surface normal at a point $\bf{p}$ as follows:
\begin{equation}\label{eq:atlas}
    \mathcal{N}_\text{atlas} = {\displaystyle\frac{\partial f_i^{\bf x}}{\partial u}} \times {\displaystyle\frac{\partial f_i^{\bf x}}{\partial v}} \bigg\rvert_{\bf p}
\end{equation}

The gradient of OccupancyNet's at a point ${\bf q}$, is computed as:
\begin{equation}\label{eq:occ}
    \mathcal{N}_\text{occ} = \nabla_{\bf{q}} g^{\bf \hat{x}}({\bf q})
\end{equation}
We now define the normal consistency loss by measuring the misalignment in their directions (note that the values are normalized to have unit magnitude):
\begin{equation}\label{eq:normal}
    \mathcal{L}_\text{norm} = \bigg|{1 - \frac{\mathcal{N}_\text{atlas}}{\norm{\mathcal{N}_\text{atlas}}} \cdot \frac{\mathcal{N}_\text{occ}}{\norm{\mathcal{N}_\text{occ}}}}\bigg|  
\end{equation}
We evaluate this loss only at surface points predicted by the explicit representation (i.e., $\mathcal{N}_\text{occ}$ is evaluated at ${\bf q}=f^{\bf x}(\bf{p}), \bf{p} \in \mathcal{A}$).

\vpara{Fitting losses.}
Each branch also has its own fitting loss, ensuring it adheres to the input geometry. We use the standard losses used to train each of the two surface representations in previous works.

For the explicit branch, we measure the distance between the predicted surface and the ground truth in standard manner, using Chamfer distance: 
\begin{equation}\label{eq:chamfer}
\mathcal{L}_\text{chamfer} 
= \sum_{{\bf p}\in \mathcal{A}} \min_{{\bf \hat{p}}\in S} | f^{\bf x}({\bf p}) - {\bf \hat{p}} |^2 + \sum_{{\bf \hat{p}}\in S} \min_{{\bf p}\in \mathcal{A}} | f^{\bf x}({\bf p}) - {\bf \hat{p}} |^2,
\end{equation}
where $\mathcal{A}$ be a set of points randomly sampled from the $K$ unit squares of the charts (here $f^{\bf x}$ uses one of the neural functions $f_i$ depending on which of the $K$ charts the point $\bf{p}$ came from). ${S}$ is a set of points that represent the ground truth surface. 

For the implicit branch, given a set of points $\{\textbf{q}_i\}_{i=1}^N$ sampled in 3D space, with $o(\textbf{q}_i)$ denoting their ground-truth occupancy values, the occupancy loss is defined as: 
\begin{equation}\label{eq:l_occ}
\mathcal{L}_\text{occ} = \sum_{i=1}^N\mathcal{H}\big(g^{\bf \hat{x}}(\textbf{q}_i), o({\textbf{q}_i}) \big)
\end{equation}

Finally, for many applications visual quality of a rendered 3D reconstruction plays a very important role (e.g., every paper on this subject actually presents a rendering of the reconstructed model for qualitative evaluations). Rendering implicit functions requires complex probing of volumes, while output of the explicit representation can be directly rasterized into an image. Thus, we chose to only include an image-space loss for the output of the explicit branch, comparing its differentiable rendering to the image produced by rendering the ground truth shape. Note that this loss to still improves the representation learned by the implicit branch due to consistency losses. 

To compute the image-space loss we first reconstruct a mesh from the explicit branch. In particular, we sample a set of 2D points $\mathcal{A}_u$ on a regular grid for each of the $K$ unit squares. Each grid defines topology of the mesh, and mapping the corners of all grids with $f^{\bf x}$ gives a triangular 3D mesh that can be used with most existing differentiable rasterizers  $\mathcal{R}$ (we use our own implementation inspired by SoftRas~\cite{liu2019soft}). We render $25$ images from different viewpoints produced by the cross product of $5$ elevation and $5$ azimuth uniformly sampled angles.

The image loss is defined as:
\begin{equation}
    \mathcal{L}_\text{img} = \frac{1}{25}\sum_{i=1}^{25} \norm{\mathcal{R}(f^{\bf x}(\mathcal{A}_u), v_i)-\mathcal{R}(\mathcal{M}_{gt}, v_i)}^2
\end{equation}
in which $v_i$ is the $i^\text{th}$ viewpoint and $\mathcal{M}_{gt}$ represents the ground truth mesh. Our renderer $\mathcal{R}$ outputs a normal map image (based on per-face normals), since they capture the shape better than silhouettes or gray-shaded images. 


Our final loss is a weighted combination of the fitting and consistency losses:
\begin{equation}\label{eq:total}
    \mathcal{L}_\text{total} = \mathcal{L}_\text{occ} + \alpha \cdot \mathcal{L}_\text{chamfer} + \beta \cdot  \mathcal{L}_\text{img} + \gamma \cdot \mathcal{L}_\text{consistency} + \delta \cdot \mathcal{L}_\text{norm}
\end{equation}
{Since the loss functions measure different quantities with vastly different scales, we set the weights empirically to get the best qualitative and quantitative results on the validation set.} 
We use $\alpha = 2.5\times 10^4$, $\beta = 10^3$, $\gamma = 0.04$, and $\delta = 0.05$ in all experiments. 

\subsection{Pipeline and Training}
Figure \ref{fig:architecture} illustrates the complete pipeline for training and inference: given an input image or a point cloud, the two encoders encode the input to shape features, ${\bf x}$ and $\hat{\bf x}$. For the AtlasNet branch, a set of points $\mathcal{A} \subset [0, 1]^2$ is randomly sampled from $K$ unit squares. These points are concatenated with the shape feature ${\bf x}$ and passed to AtlasNet. The Chamfer loss is computed between $f^{\bf x}(\mathcal{A})$ and the ground truth surface points, per Equation \ref{eq:chamfer}. For the OccupancyNet branch,
similarly to \cite{mescheder2019occupancy}, we uniformly sample a set of points $\{q_i\}_{i=1}^{N}\subset \mathbb{R}^3$ inside the bounding box of the object 
and use them to train OccupancyNet with respect to the fitting losses. 
To compute the image loss, the generated mesh $f^{\bf x}(\mathcal{A}_{u})$ and the ground truth mesh $\mathcal{M}_{gt}$ are normalized to a unit cube prior to rendering.

For the consistency loss, the occupancy function $g^{\bf \hat{x}}$ is evaluated at the points generated by AtlasNet, $f^{\bf x}(\mathcal{A})$ and then penalized as described in Equation \eqref{eq:consistency}. 
AtlasNet's normals are evaluated at the sample points $\mathcal{A}$. OccupancyNet's normals are evaluated at the corresponding points, $f^{\bf x}(\mathcal{A})$. These are then plugged into the loss described in Equation \eqref{eq:normal}. 
We train AtlasNet and OccupancyNet jointly with the loss function in equation \eqref{eq:total}, thereby coupling the two branches to one-another via the consistency losses.  

Since we wish to show the merit of the hybrid approach, we keep the two branches' networks' architecture and training setup identical to the one used in the previous works that introduced those two networks. For AtlasNet, we sample random $100$ points from each of $K=25$ patches during training. At inference time, the points are sampled on $10 \times 10$ regular grid for each patch. For OccupancyNet, we use the $2500$ uniform samples provided by the authors~\cite{mescheder2019occupancy}. We use the Adam optimizer \cite{kingma2014adam} with learning rates of $6 \times 10^{-4}$ and $1.5 \times 10^{-4}$ for AtlasNet ($f$) and OccupancyNet ($g$) respectively.

\section{Results}
\begin{table}[!t]
\begin{center}
\begin{tabu}{ |[1.5pt]c|[1pt]c|c|c|c|p{0.7cm}|p{0.7cm}|p{0.9cm}|p{0.9cm}|p{0.9cm}|p{0.9cm}|p{1.3cm}|p{1.3cm}|[1.5pt] }
 \tabucline[1.5pt]{-}
 Metric & \multicolumn{12}{c|[1.5pt]}{\bf Chamfer-$L_1 (\times 10^{-1})$} \\
 \tabucline[1pt]{-} 
  Model & \multicolumn{2}{c|}{AN} & \multicolumn{2}{c|}{ ON} & \multicolumn{2}{c|}{Hybrid} & \multicolumn{2}{c|}{
  No $\mathcal{L}_\text{img}$ } & \multicolumn{2}{c|}{
  No $\mathcal{L}_\text{norm}$ } & \multicolumn{2}{c|[1.5pt]}{
  No $\mathcal{L}_\text{img}, \mathcal{L}_\text{norm}$ } \\
   \hline
  Branch & \multicolumn{2}{c|}{} & \multicolumn{2}{c|}{} & \multicolumn{1}{c|}{AN} & \multicolumn{1}{c|}{ON} & \multicolumn{1}{c|}{AN} & \multicolumn{1}{c|}{ON} & \multicolumn{1}{c|}{AN} & \multicolumn{1}{c|}{ON} & \multicolumn{1}{c|}{AN} & \multicolumn{1}{c|[1.5pt]}{ON} \\
 \tabucline[1pt]{-}
 airplane &  \multicolumn{2}{c|}{1.05} & \multicolumn{2}{c|}{1.34} & \multicolumn{1}{c|}{\bf 0.91} & \multicolumn{1}{c|}{1.03} & \multicolumn{1}{c|}{0.96} & \multicolumn{1}{c|}{1.10} & \multicolumn{1}{c|}{0.95} & \multicolumn{1}{c|}{1.08} & \multicolumn{1}{c|}{1.01} & \multicolumn{1}{c|[1.5pt]}{1.17} \\
 bench & \multicolumn{2}{c|}{1.38} & \multicolumn{2}{c|}{1.50} & \multicolumn{1}{c|}{\bf 1.23} & \multicolumn{1}{c|}{1.26} & \multicolumn{1}{c|}{1.27} & \multicolumn{1}{c|}{1.31} & \multicolumn{1}{c|}{1.26} & \multicolumn{1}{c|}{1.29} & \multicolumn{1}{c|}{1.32} & \multicolumn{1}{c|[1.5pt]}{1.38} \\
 cabinet & \multicolumn{2}{c|}{1.75} & \multicolumn{2}{c|}{1.53} & \multicolumn{1}{c|}{1.53} & \multicolumn{1}{c|}{\bf 1.47} & \multicolumn{1}{c|}{1.57} & \multicolumn{1}{c|}{1.49} & \multicolumn{1}{c|}{1.55} & \multicolumn{1}{c|}{1.49} & \multicolumn{1}{c|}{1.61} & \multicolumn{1}{c|[1.5pt]}{1.50} \\
 car & \multicolumn{2}{c|}{1.41} & \multicolumn{2}{c|}{1.49} & \multicolumn{1}{c|}{\bf 1.28} & \multicolumn{1}{c|}{1.31} & \multicolumn{1}{c|}{1.33} & \multicolumn{1}{c|}{1.37} & \multicolumn{1}{c|}{1.33} & \multicolumn{1}{c|}{1.36} & \multicolumn{1}{c|}{1.37} & \multicolumn{1}{c|[1.5pt]}{1.42} \\
 chair & \multicolumn{2}{c|}{2.09} & \multicolumn{2}{c|}{2.06} & \multicolumn{1}{c|}{1.96}  & \multicolumn{1}{c|}{\bf 1.95} & \multicolumn{1}{c|}{2.02} & \multicolumn{1}{c|}{2.01} & \multicolumn{1}{c|}{1.99} & \multicolumn{1}{c|}{1.99} & \multicolumn{1}{c|}{2.04} & \multicolumn{1}{c|[1.5pt]}{2.03} \\
 display & \multicolumn{2}{c|}{1.98} & \multicolumn{2}{c|}{2.58} & \multicolumn{1}{c|}{\bf 1.89} & \multicolumn{1}{c|}{2.14}  & \multicolumn{1}{c|}{1.92} & \multicolumn{1}{c|}{2.24} & \multicolumn{1}{c|}{1.90} & \multicolumn{1}{c|}{2.19} & \multicolumn{1}{c|}{1.94} & \multicolumn{1}{c|[1.5pt]}{2.29} \\
 lamp & \multicolumn{2}{c|}{3.05} & \multicolumn{2}{c|}{3.68} & \multicolumn{1}{c|}{\bf 2.91} & \multicolumn{1}{c|}{3.02} & \multicolumn{1}{c|}{2.93} & \multicolumn{1}{c|}{3.09} & \multicolumn{1}{c|}{2.91} & \multicolumn{1}{c|}{3.06} & \multicolumn{1}{c|}{2.99} & \multicolumn{1}{c|[1.5pt]}{3.21} \\
 sofa & \multicolumn{2}{c|}{1.77} & \multicolumn{2}{c|}{1.81} & \multicolumn{1}{c|}{\bf 1.56} & \multicolumn{1}{c|}{1.58} & \multicolumn{1}{c|}{1.61} & \multicolumn{1}{c|}{1.63} & \multicolumn{1}{c|}{1.59} & \multicolumn{1}{c|}{1.61} & \multicolumn{1}{c|}{1.68} & \multicolumn{1}{c|[1.5pt]}{1.71} \\
 table & \multicolumn{2}{c|}{1.90} & \multicolumn{2}{c|}{1.82} & \multicolumn{1}{c|}{1.73} & \multicolumn{1}{c|}{\bf 1.72} & \multicolumn{1}{c|}{1.80} & \multicolumn{1}{c|}{1.78} & \multicolumn{1}{c|}{1.78} & \multicolumn{1}{c|}{1.76} & \multicolumn{1}{c|}{1.83} & \multicolumn{1}{c|[1.5pt]}{1.79} \\
 telephone & \multicolumn{2}{c|}{1.28} & \multicolumn{2}{c|}{1.27} & \multicolumn{1}{c|}{\bf 1.17} & \multicolumn{1}{c|}{1.18} & \multicolumn{1}{c|}{1.22} & \multicolumn{1}{c|}{1.21} & \multicolumn{1}{c|}{1.19} & \multicolumn{1}{c|}{1.19} & \multicolumn{1}{c|}{1.24} & \multicolumn{1}{c|[1.5pt]}{1.24} \\
 vessel & \multicolumn{2}{c|}{1.51} & \multicolumn{2}{c|}{2.01} & \multicolumn{1}{c|}{\bf 1.42} & \multicolumn{1}{c|}{1.53} & \multicolumn{1}{c|}{1.46} & \multicolumn{1}{c|}{1.60} & \multicolumn{1}{c|}{1.46} & \multicolumn{1}{c|}{1.58} & \multicolumn{1}{c|}{1.48} & \multicolumn{1}{c|[1.5pt]}{1.69} \\
 \hline
 mean & \multicolumn{2}{c|}{1.74} & \multicolumn{2}{c|}{1.92} & \multicolumn{1}{c|}{\bf 1.60} & \multicolumn{1}{c|}{1.65} 
& \multicolumn{1}{c|}{1.64} &\multicolumn{1}{c|}{1.71}  & \multicolumn{1}{c|}{1.63} & \multicolumn{1}{c|}{1.69} & \multicolumn{1}{c|}{1.68} & \multicolumn{1}{c|[1.5pt]}{1.77}  \\
\tabucline[1.5pt]{-}
 Metric & \multicolumn{12}{c|[1.5pt]}{\bf Normal Consistency ($\times 10^{-2}$)}  \\
 \tabucline[1pt]{-} 
  Model & \multicolumn{2}{c|}{AN} & \multicolumn{2}{c|}{ON} & \multicolumn{2}{c|}{Hybrid} & \multicolumn{2}{c|}{No $\mathcal{L}_\text{img}$} & \multicolumn{2}{c|}{No $\mathcal{L}_\text{norm}$} & \multicolumn{2}{c|[1.5pt]}{No $\mathcal{L}_\text{img},\mathcal{L}_\text{norm}$} \\
   \hline
  Branch & \multicolumn{2}{c|}{} & \multicolumn{2}{c|}{} & \multicolumn{1}{c|}{AN} & \multicolumn{1}{c|}{ON} & \multicolumn{1}{c|}{AN} & \multicolumn{1}{c|}{ON} & \multicolumn{1}{c|}{AN} & \multicolumn{1}{c|}{ON} & \multicolumn{1}{c|}{AN} & \multicolumn{1}{c|[1.5pt]}{ON} \\
 \tabucline[1pt]{-}
 airplane &  \multicolumn{2}{c|}{83.6} & \multicolumn{2}{c|}{84.5} & \multicolumn{1}{c|}{85.5} & \multicolumn{1}{c|}{\bf 85.7} & \multicolumn{1}{c|}{85.3} & \multicolumn{1}{c|}{85.6} & \multicolumn{1}{c|}{84.8} & \multicolumn{1}{c|}{85.3} & \multicolumn{1}{c|}{84.3} & \multicolumn{1}{c|[1.5pt]}{85.0}\\
 bench & \multicolumn{2}{c|}{77.9} & \multicolumn{2}{c|}{81.4} & \multicolumn{1}{c|}{81.4} & \multicolumn{1}{c|}{\bf 82.5} & \multicolumn{1}{c|}{80.9} & \multicolumn{1}{c|}{82.2} & \multicolumn{1}{c|}{80.4} & \multicolumn{1}{c|}{81.9} & \multicolumn{1}{c|}{79.9} & \multicolumn{1}{c|[1.5pt]}{81.7} \\
 cabinet & \multicolumn{2}{c|}{85.0} & \multicolumn{2}{c|}{88.4} & \multicolumn{1}{c|}{88.3} & \multicolumn{1}{c|}{\bf 89.1} & \multicolumn{1}{c|}{88.1} & \multicolumn{1}{c|}{89.0} & \multicolumn{1}{c|}{87.2} & \multicolumn{1}{c|}{88.7} & \multicolumn{1}{c|}{86.8} & \multicolumn{1}{c|[1.5pt]}{88.6} \\
 car & \multicolumn{2}{c|}{83.6} & \multicolumn{2}{c|}{85.2} & \multicolumn{1}{c|}{86.2} & \multicolumn{1}{c|}{\bf 86.8} & \multicolumn{1}{c|}{85.8} & \multicolumn{1}{c|}{86.5} & \multicolumn{1}{c|}{85.3} & \multicolumn{1}{c|}{86.0} & \multicolumn{1}{c|}{84.9} & \multicolumn{1}{c|[1.5pt]}{85.8} \\
 chair & \multicolumn{2}{c|}{79.1} & \multicolumn{2}{c|}{82.9} & \multicolumn{1}{c|}{83.5} & \multicolumn{1}{c|}{\bf 84.0} & \multicolumn{1}{c|}{83.1} & \multicolumn{1}{c|}{83.7} & \multicolumn{1}{c|}{82.4} & \multicolumn{1}{c|}{83.4} & \multicolumn{1}{c|}{82.0} & \multicolumn{1}{c|[1.5pt]}{83.2} \\
 display & \multicolumn{2}{c|}{85.8} & \multicolumn{2}{c|}{85.7} & \multicolumn{1}{c|}{\bf 87.0} & \multicolumn{1}{c|}{86.9} & \multicolumn{1}{c|}{86.7} & \multicolumn{1}{c|}{86.6} & \multicolumn{1}{c|}{86.3} & \multicolumn{1}{c|}{86.1} & \multicolumn{1}{c|}{86.0} & \multicolumn{1}{c|[1.5pt]}{85.9} \\
 lamp & \multicolumn{2}{c|}{69.4} & \multicolumn{2}{c|}{75.1} & \multicolumn{1}{c|}{74.9} & \multicolumn{1}{c|}{\bf 76.0} & \multicolumn{1}{c|}{74.7} & \multicolumn{1}{c|}{75.9} & \multicolumn{1}{c|}{73.3} & \multicolumn{1}{c|}{75.6} & \multicolumn{1}{c|}{72.8} & \multicolumn{1}{c|[1.5pt]}{75.4} \\
 sofa & \multicolumn{2}{c|}{84.0} & \multicolumn{2}{c|}{86.7} & \multicolumn{1}{c|}{87.2} & \multicolumn{1}{c|}{\bf 87.5} & \multicolumn{1}{c|}{86.9} & \multicolumn{1}{c|}{87.4} & \multicolumn{1}{c|}{86.4} & \multicolumn{1}{c|}{87.1} & \multicolumn{1}{c|}{85.9} & \multicolumn{1}{c|[1.5pt]}{86.9} \\
 table & \multicolumn{2}{c|}{83.2} & \multicolumn{2}{c|}{85.8} & \multicolumn{1}{c|}{86.3} & \multicolumn{1}{c|}{\bf 87.4} & \multicolumn{1}{c|}{86.0} & \multicolumn{1}{c|}{87.1} & \multicolumn{1}{c|}{85.3} & \multicolumn{1}{c|}{86.4} & \multicolumn{1}{c|}{84.9} & \multicolumn{1}{c|[1.5pt]}{86.1} \\
 telephone & \multicolumn{2}{c|}{92.3} & \multicolumn{2}{c|}{93.9} & \multicolumn{1}{c|}{94.0} & \multicolumn{1}{c|}{\bf 94.5}  & \multicolumn{1}{c|}{93.8} & \multicolumn{1}{c|}{94.4} & \multicolumn{1}{c|}{93.6} & \multicolumn{1}{c|}{94.2} & \multicolumn{1}{c|}{93.3} & \multicolumn{1}{c|[1.5pt]}{94.1} \\
 vessel & \multicolumn{2}{c|}{75.6} & \multicolumn{2}{c|}{79.7} &  \multicolumn{1}{c|}{79.2} & \multicolumn{1}{c|}{\bf 80.6} & \multicolumn{1}{c|}{78.9} & \multicolumn{1}{c|}{80.4} & \multicolumn{1}{c|}{77.7} & \multicolumn{1}{c|}{80.0} & \multicolumn{1}{c|}{77.4} & \multicolumn{1}{c|[1.5pt]}{79.9} \\
 \hline
 mean & \multicolumn{2}{c|}{81.8} & \multicolumn{2}{c|}{84.5} & \multicolumn{1}{c|}{84.9} & \multicolumn{1}{c|}{\bf 85.5} 
& \multicolumn{1}{c|}{84.6} & \multicolumn{1}{c|}{85.4} & \multicolumn{1}{c|}{83.9} & \multicolumn{1}{c|}{85.0} & \multicolumn{1}{c|}{83.5} & \multicolumn{1}{c|[1.5pt]}{84.8} \\
\tabucline[1.5pt]{-}
\end{tabu}
\end{center}
\caption{Quantitative results on single-view reconstruction. Variants of our hybrid model, with AtlasNet (AN) and OccupancyNet (ON) branches, are compared with vanilla AtlasNet and OccupancyNet using Chamfer-$L_1$ distance and Normal Consistenty score.  
} \label{tab:quant-svr}
\end{table}

\begin{figure}[t]
\begin{center}
   \includegraphics[width=1.0\linewidth]{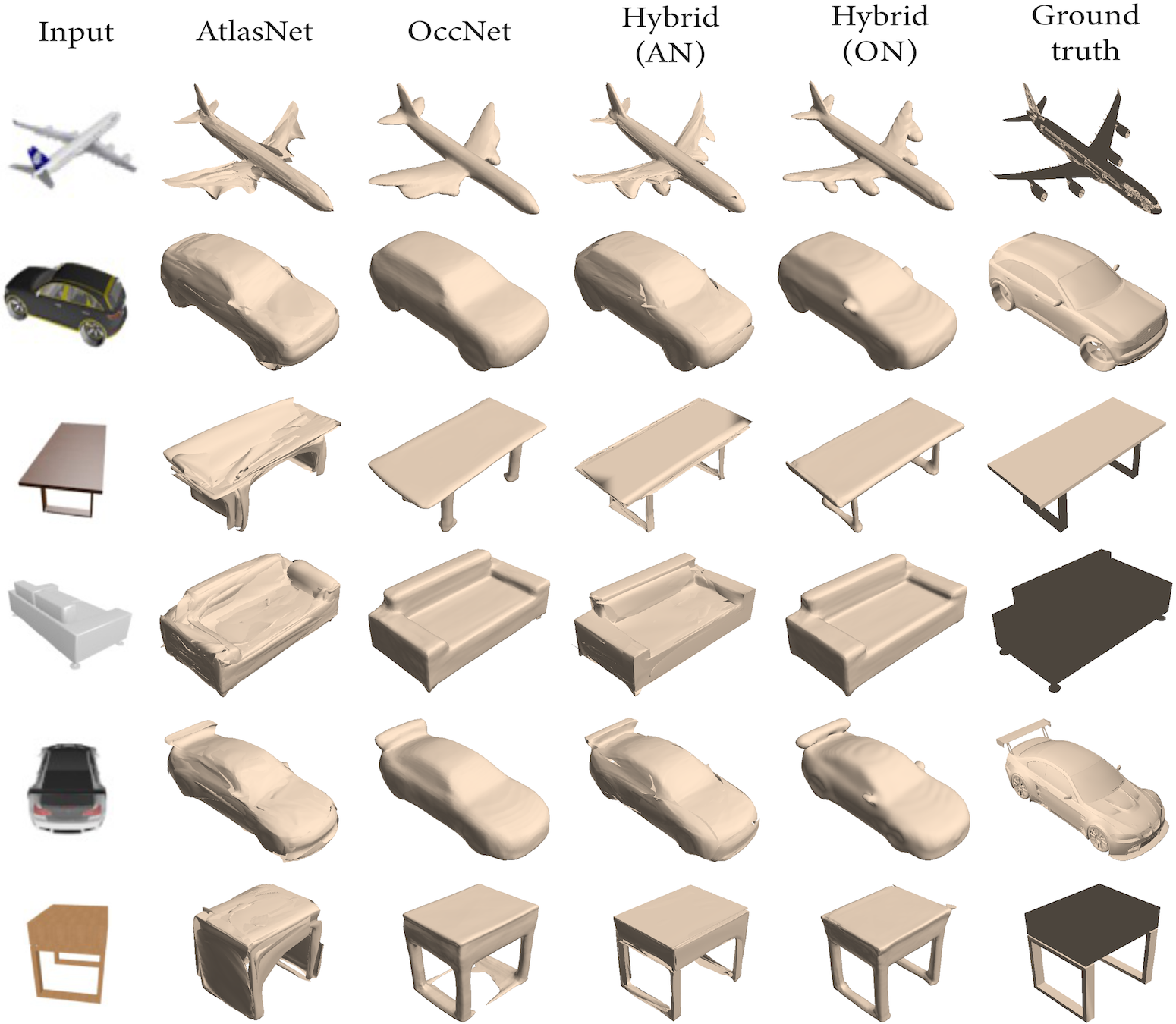}
\end{center}
   \caption{Comparison of meshes generated by vanilla AtlasNet and OccupancyNet with the AtlasNet (AN) and OccupancyNet (ON) branches of our hybrid model. Compared to their vanilla counterparts, our AtlasNet branch produces results with significantly less oscillatory artifacts, and our OccupancyNet branch produces results that better preserve thin features such as the chair legs.  }\label{fig:qual-img}
\vspace{-0.2cm}
\end{figure}

We evaluate our network's performance on single view reconstruction as well as on point cloud reconstruction, using the same subset of shapes from ShapeNet \cite{chang2015shapenet} as  used in Choy et al. \cite{choy20163d}. For both tasks, following prior work (e.g., ~\cite{groueix2018atlasnet,mescheder2019occupancy}), we use simple encoder-decoder architectures. 
Similarly to \cite{mescheder2019occupancy}, we quantitatively evaluate the results using the chamfer-$L_1$ distance and normal consistency score.  
The chamfer-$L_1$ distance is the mean of the accuracy and completeness metrics, with accuracy being the average distance of points on the output mesh to their nearest neighbors on the ground truth mesh, and completeness similarly with switching the roles of source and target point sets. 
{Note that we use the chamfer-$L_2$ distance for training in order to be consistent with the AtlasNet paper~\cite{groueix2018atlasnet}. For evaluation, we use the chamfer-$L_1$ distance since it is adopted as the evaluation metric in OccupancyNet~\cite{mescheder2019occupancy}.}
The normal consistency score is the mean absolute dot product of normals in the predicted surface and normals at the corresponding nearest neighbors on the true surface.

\vpara{Single view reconstruction.}
To reconstruct geometry from a single-view image, we use a ResNet-18 \cite{he2016deep} encoder for each of the two branches to encode an input image into a shape descriptor which is then fed to the branch. 
{Using distinct encoders enables model-specific feature extraction, and we found this to slightly outperform a shared encoder.} 
We then train end-to-end with the loss \eqref{eq:total}, on the dataset of images provided by Choy et al.~\cite{choy20163d}, using batch size of $7$. Note that with our method the surface can be reconstructed from either the explicit AtlasNet (AN) branch or the implicit OccupancyNet (ON) branch. We show qualitative results (Figure~\ref{fig:qual-img}) and error metrics (Table~\ref{tab:quant-svr}) for both branches. 
The surface generated by our AtlasNet branch, ``Hybrid (AN),'' provides a visually smoother surface than  vanilla AtlasNet (AN), which is also closer to the ground truth -- both in terms of chamfer distance, as well as its normal-consistency score. The surface generated by our OccupancyNet branch, ``Hybrid (ON)'', similarly yields a more accurate surface in comparison to vanilla OccupancyNet (ON). We observe that the hybrid implicit representation tends to be better at capturing thinner surfaces (e.g., see table legs in Figure~\ref{fig:qual-img}) than its vanilla counterpart; this improvement is exactly due to having the implicit branch indirectly trained with the chamfer loss propagated from the AtlasNet branch.

\begin{table}[!tb]
\begin{center}
\begin{tabu}{ |[1.5pt]c|[1pt]c|c|c|c|p{0.7cm}|p{0.7cm}|p{0.9cm}|p{0.9cm}|p{0.9cm}|p{0.9cm}|p{1.3cm}|p{1.3cm}|[1.5pt] }
 \tabucline[1.5pt]{-}
 Metric & \multicolumn{12}{c|[1.5pt]}{\bf Chamfer-$L_1 (\times 10^{-3})$} \\
 \tabucline[1pt]{-} 
  Model & \multicolumn{2}{c|}{AN} & \multicolumn{2}{c|}{ON} & \multicolumn{2}{c|}{Hybrid} & \multicolumn{2}{c|}{No $\mathcal{L}_\text{img}$} & \multicolumn{2}{c|}{No $\mathcal{L}_\text{norm}$} & \multicolumn{2}{c|[1.5pt]}{No $\mathcal{L}_\text{img}, \mathcal{L}_\text{norm}$} \\
  \hline
  Branch & \multicolumn{2}{c|}{} & \multicolumn{2}{c|}{} & \multicolumn{1}{c|}{AN} & \multicolumn{1}{c|}{ON} & \multicolumn{1}{c|}{AN} & \multicolumn{1}{c|}{ON} & \multicolumn{1}{c|}{AN} & \multicolumn{1}{c|}{ON} & \multicolumn{1}{c|}{AN} & \multicolumn{1}{c|[1.5pt]}{ON} \\
 \tabucline[1pt]{-}
 airplane & \multicolumn{2}{c|}{0.17} & \multicolumn{2}{c|}{0.19} & \multicolumn{1}{c|}{\bf 0.15} & \multicolumn{1}{c|}{0.16} & \multicolumn{1}{c|}{0.16} & \multicolumn{1}{c|}{0.17} & \multicolumn{1}{c|}{0.16} & \multicolumn{1}{c|}{0.17} & \multicolumn{1}{c|}{0.17} & \multicolumn{1}{c|[1.5pt]}{0.18} \\
 bench & \multicolumn{2}{c|}{0.49} & \multicolumn{2}{c|}{\bf 0.23} & \multicolumn{1}{c|}{0.31} & \multicolumn{1}{c|}{0.25} & \multicolumn{1}{c|}{0.34} & \multicolumn{1}{c|}{0.25} & \multicolumn{1}{c|}{0.33} & \multicolumn{1}{c|}{0.24} & \multicolumn{1}{c|}{0.37} & \multicolumn{1}{c|[1.5pt]}{0.24} \\
 cabinet & \multicolumn{2}{c|}{0.73} & \multicolumn{2}{c|}{0.56} & \multicolumn{1}{c|}{0.55} &\multicolumn{1}{c|}{\bf 0.51} & \multicolumn{1}{c|}{0.58} & \multicolumn{1}{c|}{0.52} & \multicolumn{1}{c|}{0.61} & \multicolumn{1}{c|}{0.54} & \multicolumn{1}{c|}{0.63} & \multicolumn{1}{c|[1.5pt]}{0.54} \\
 car & \multicolumn{2}{c|}{0.49} & \multicolumn{2}{c|}{0.54} & \multicolumn{1}{c|}{\bf 0.42} & \multicolumn{1}{c|}{0.44} & \multicolumn{1}{c|}{0.46} & \multicolumn{1}{c|}{0.47} & \multicolumn{1}{c|}{0.44} & \multicolumn{1}{c|}{0.47} & \multicolumn{1}{c|}{0.47} & \multicolumn{1}{c|[1.5pt]}{0.50} \\
 chair & \multicolumn{2}{c|}{0.52} & \multicolumn{2}{c|}{\bf 0.30} & \multicolumn{1}{c|}{0.36} & \multicolumn{1}{c|}{0.33} & \multicolumn{1}{c|}{0.39} & \multicolumn{1}{c|}{0.33} & \multicolumn{1}{c|}{0.38} & \multicolumn{1}{c|}{0.33} & \multicolumn{1}{c|}{0.41} & \multicolumn{1}{c|[1.5pt]}{0.32} \\
 display & \multicolumn{2}{c|}{0.61} & \multicolumn{2}{c|}{0.45} & \multicolumn{1}{c|}{0.47} & \multicolumn{1}{c|}{\bf 0.39} & \multicolumn{1}{c|}{0.50} & \multicolumn{1}{c|}{0.41} & \multicolumn{1}{c|}{0.48} & \multicolumn{1}{c|}{0.40} & \multicolumn{1}{c|}{0.52} & \multicolumn{1}{c|[1.5pt]}{0.42} \\
 lamp & \multicolumn{2}{c|}{1.53} & \multicolumn{2}{c|}{1.35} & \multicolumn{1}{c|}{1.42} & \multicolumn{1}{c|}{\bf 1.31} & \multicolumn{1}{c|}{1.46} & \multicolumn{1}{c|}{1.33} & \multicolumn{1}{c|}{1.44} & \multicolumn{1}{c|}{1.31} & \multicolumn{1}{c|}{1.49} & \multicolumn{1}{c|[1.5pt]}{1.34} \\
 sofa & \multicolumn{2}{c|}{0.32} & \multicolumn{2}{c|}{0.34} & \multicolumn{1}{c|}{\bf 0.25} & \multicolumn{1}{c|}{0.26} & \multicolumn{1}{c|}{0.28} & \multicolumn{1}{c|}{0.29} & \multicolumn{1}{c|}{0.26} & \multicolumn{1}{c|}{0.27} & \multicolumn{1}{c|}{0.30} & \multicolumn{1}{c|[1.5pt]}{0.31} \\
 table & \multicolumn{2}{c|}{0.58} & \multicolumn{2}{c|}{0.45} & \multicolumn{1}{c|}{0.46} & \multicolumn{1}{c|}{\bf 0.41} & \multicolumn{1}{c|}{0.48} & \multicolumn{1}{c|}{0.42} & \multicolumn{1}{c|}{0.47} & \multicolumn{1}{c|}{0.42} & \multicolumn{1}{c|}{0.50} & \multicolumn{1}{c|[1.5pt]}{0.43} \\
 telephone & \multicolumn{2}{c|}{0.22} & \multicolumn{2}{c|}{0.12} & \multicolumn{1}{c|}{0.14} & \multicolumn{1}{c|}{\bf 0.10} & \multicolumn{1}{c|}{0.15} & \multicolumn{1}{c|}{0.10} & \multicolumn{1}{c|}{0.16} & \multicolumn{1}{c|}{0.11} & \multicolumn{1}{c|}{0.18} & \multicolumn{1}{c|[1.5pt]}{0.11} \\
 watercraft & \multicolumn{2}{c|}{0.74} & \multicolumn{2}{c|}{\bf 0.38} & \multicolumn{1}{c|}{0.53} & \multicolumn{1}{c|}{0.42} & \multicolumn{1}{c|}{0.57} & \multicolumn{1}{c|}{0.41} & \multicolumn{1}{c|}{0.54} & \multicolumn{1}{c|}{0.42} & \multicolumn{1}{c|}{0.61} & \multicolumn{1}{c|[1.5pt]}{0.40} \\
 \hline
 mean & \multicolumn{2}{c|}{0.58} & \multicolumn{2}{c|}{0.45} & \multicolumn{1}{c|}{0.46} & \multicolumn{1}{c|}{\bf 0.41} & \multicolumn{1}{c|}{0.49} & \multicolumn{1}{c|}{0.43} & \multicolumn{1}{c|}{0.48} & \multicolumn{1}{c|}{0.43} & \multicolumn{1}{c|}{0.51} & \multicolumn{1}{c|[1.5pt]}{0.44} \\
\tabucline[1.5pt]{-}
 Metric & \multicolumn{12}{c|[1.5pt]}{\bf Normal Consistency ($\times 10^{-2}$)}  \\
 \tabucline[1pt]{-} 
  Model & \multicolumn{2}{c|}{AN} & \multicolumn{2}{c|}{ON} & \multicolumn{2}{c|}{Hybrid} & \multicolumn{2}{c|}{No $\mathcal{L}_\text{img}$} & \multicolumn{2}{c|}{No $\mathcal{L}_\text{norm}$} & \multicolumn{2}{c|[1.5pt]}{No $\mathcal{L}_\text{img}, \mathcal{L}_\text{norm}$} \\
  \hline
  Branch & \multicolumn{2}{c|}{} & \multicolumn{2}{c|}{} & \multicolumn{1}{c|}{AN} & \multicolumn{1}{c|}{ON} & \multicolumn{1}{c|}{AN} & \multicolumn{1}{c|}{ON} & \multicolumn{1}{c|}{AN} & \multicolumn{1}{c|}{ON} & \multicolumn{1}{c|}{AN} & \multicolumn{1}{c|[1.5pt]}{ON} \\
 \tabucline[1pt]{-}
 airplane &  \multicolumn{2}{c|}{85.4} & \multicolumn{2}{c|}{89.6} & \multicolumn{1}{c|}{88.3} & \multicolumn{1}{c|}{\bf 90.1} & \multicolumn{1}{c|}{88.1} & \multicolumn{1}{c|}{90.0} & \multicolumn{1}{c|}{87.5} & \multicolumn{1}{c|}{89.7} & \multicolumn{1}{c|}{87.1} & \multicolumn{1}{c|[1.5pt]}{89.6} \\
 bench & \multicolumn{2}{c|}{81.5} & \multicolumn{2}{c|}{\bf 87.1} & \multicolumn{1}{c|}{85.6} & \multicolumn{1}{c|}{86.7} & \multicolumn{1}{c|}{85.3} & \multicolumn{1}{c|}{86.8} & \multicolumn{1}{c|}{85.0} & \multicolumn{1}{c|}{86.8} & \multicolumn{1}{c|}{84.7} & \multicolumn{1}{c|[1.5pt]}{86.9} \\
 cabinet & \multicolumn{2}{c|}{87.0} & \multicolumn{2}{c|}{90.6} & \multicolumn{1}{c|}{89.3} & \multicolumn{1}{c|}{\bf 91.1} & \multicolumn{1}{c|}{89.1} & \multicolumn{1}{c|}{91.0} & \multicolumn{1}{c|}{88.4} & \multicolumn{1}{c|}{90.8} & \multicolumn{1}{c|}{88.1} & \multicolumn{1}{c|[1.5pt]}{90.8} \\
 car & \multicolumn{2}{c|}{84.7} & \multicolumn{2}{c|}{87.9} & \multicolumn{1}{c|}{87.5} & \multicolumn{1}{c|}{\bf 88.6} & \multicolumn{1}{c|}{87.1} & \multicolumn{1}{c|}{88.5} & \multicolumn{1}{c|}{86.7} & \multicolumn{1}{c|}{88.1} & \multicolumn{1}{c|}{86.1} & \multicolumn{1}{c|[1.5pt]}{88.0} \\
 chair & \multicolumn{2}{c|}{84.7} & \multicolumn{2}{c|}{\bf 94.9} & \multicolumn{1}{c|}{88.7} & \multicolumn{1}{c|}{94.3} & \multicolumn{1}{c|}{88.2} & \multicolumn{1}{c|}{94.5} & \multicolumn{1}{c|}{87.6} & \multicolumn{1}{c|}{94.5} & \multicolumn{1}{c|}{87.1} & \multicolumn{1}{c|[1.5pt]}{94.6} \\
 display & \multicolumn{2}{c|}{89.7} & \multicolumn{2}{c|}{91.9} & \multicolumn{1}{c|}{91.8} & \multicolumn{1}{c|}{\bf 92.4} & \multicolumn{1}{c|}{91.5} & \multicolumn{1}{c|}{92.3} & \multicolumn{1}{c|}{91.0} & \multicolumn{1}{c|}{92.2} & \multicolumn{1}{c|}{90.8} & \multicolumn{1}{c|[1.5pt]}{92.1} \\
 lamp & \multicolumn{2}{c|}{73.1} & \multicolumn{2}{c|}{79.5} & \multicolumn{1}{c|}{77.1} & \multicolumn{1}{c|}{\bf 79.8} & \multicolumn{1}{c|}{76.9} & \multicolumn{1}{c|}{79.7} & \multicolumn{1}{c|}{76.4} & \multicolumn{1}{c|}{79.6} & \multicolumn{1}{c|}{76.0} & \multicolumn{1}{c|[1.5pt]}{79.5} \\
 sofa & \multicolumn{2}{c|}{89.1} & \multicolumn{2}{c|}{92.2} & \multicolumn{1}{c|}{91.8} & \multicolumn{1}{c|}{\bf 92.8} & \multicolumn{1}{c|}{91.6} & \multicolumn{1}{c|}{92.7} & \multicolumn{1}{c|}{91.3} & \multicolumn{1}{c|}{92.5} & \multicolumn{1}{c|}{91.0} & \multicolumn{1}{c|[1.5pt]}{92.4} \\
 table & \multicolumn{2}{c|}{86.3} & \multicolumn{2}{c|}{91.0} & \multicolumn{1}{c|}{88.8} & \multicolumn{1}{c|}{\bf 91.4} & \multicolumn{1}{c|}{88.6} & \multicolumn{1}{c|}{91.3} & \multicolumn{1}{c|}{88.3} & \multicolumn{1}{c|}{91.3} & \multicolumn{1}{c|}{88.0} & \multicolumn{1}{c|[1.5pt]}{91.2} \\
 telephone & \multicolumn{2}{c|}{95.9} & \multicolumn{2}{c|}{97.3} & \multicolumn{1}{c|}{97.4} & \multicolumn{1}{c|}{\bf 98.0} & \multicolumn{1}{c|}{97.2} & \multicolumn{1}{c|}{97.8} & \multicolumn{1}{c|}{96.8} & \multicolumn{1}{c|}{97.6} & \multicolumn{1}{c|}{96.5} & \multicolumn{1}{c|[1.5pt]}{97.5} \\
 watercraft & \multicolumn{2}{c|}{82.1} & \multicolumn{2}{c|}{\bf 86.7} & \multicolumn{1}{c|}{84.9} & \multicolumn{1}{c|}{86.5} & \multicolumn{1}{c|}{84.7} & \multicolumn{1}{c|}{86.5} & \multicolumn{1}{c|}{84.3} & \multicolumn{1}{c|}{86.7} & \multicolumn{1}{c|}{84.0} & \multicolumn{1}{c|[1.5pt]}{86.7} \\
 \hline
 mean & \multicolumn{2}{c|}{85.4} & \multicolumn{2}{c|}{89.8} & \multicolumn{1}{c|}{88.3} & \multicolumn{1}{c|}{\bf 90.2} & \multicolumn{1}{c|}{88.0} & \multicolumn{1}{c|}{90.1} & \multicolumn{1}{c|}{87.6} & \multicolumn{1}{c|}{90.0} & \multicolumn{1}{c|}{87.2} & \multicolumn{1}{c|[1.5pt]}{89.9} \\
\tabucline[1.5pt]{-}
\end{tabu}
\end{center}
\caption{Quantitative results on auto-encoding. Variants of our hybrid model are compared with vanilla AtlasNet and OccupancyNet.  
} \label{tab:quant-ae}
\end{table}

\begin{figure}[t]
\begin{center}
   \includegraphics[width=1.0\linewidth]{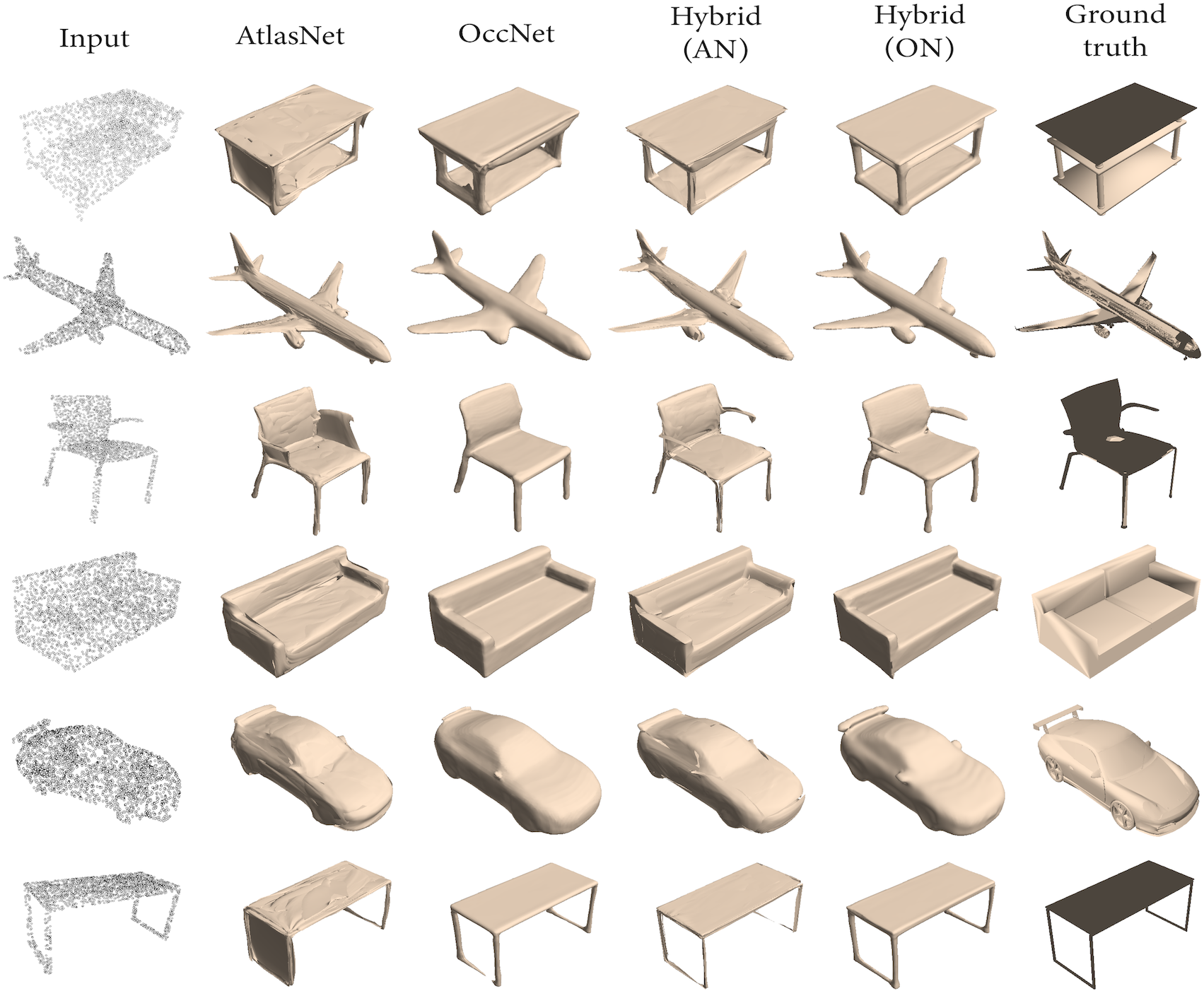}
\end{center}
\vspace{-0.35cm}
   \caption{Reconstructing surfaces from point clouds. Our hybrid approach better reproduces fine features and avoids oscillatory surface normals.}\label{fig:qual-pc}
\vspace{-0.2cm}
\end{figure}

\vpara{Point cloud reconstruction.}
As a second application, we train our network to reconstruct a surface for a sparse set of $2500$ input points. We encode the set of points to a shape descriptor using 
a PointNet~\cite{qi2017pointnet} 
encoder for each of the two branches, and train the encoder-decoder architecture end-to-end with the loss \eqref{eq:total}. We train with the same points as~\cite{mescheder2019occupancy} with a batch size of $10$. See results in Figure~\ref{fig:qual-pc} and Table~\ref{tab:quant-ae}. As in the single view reconstruction task, the hybrid method surpasses  the vanilla, single-branch architectures on average. While there are three categories in which vanilla OccupancyNet performs better, we note that Hybrid AtlasNet consistently outperforms vanilla AtlasNet on all categories. This indicates that the hybrid training is mostly beneficial for the implicit representation, and \emph{always} beneficial for the explicit representation; this in turn offers a more streamlined surface reconstruction process. 
{Additional examples are shown in the supplementary material.} 

\vpara{Ablation study on the loss functions.}
We evaluate the importance of the different loss terms via an ablation study for both tasks (see Tables~\ref{tab:quant-svr},~\ref{tab:quant-ae}). First, we exclude the image-based loss function $\mathcal{L}_\text{img}$. Note that even without this loss, hybrid AtlasNet still outperforms vanilla AtlasNet, attributing these improvements mainly to the consistency losses. Removing the normal-consistency loss $\mathcal{L}_\text{norm}$ results in decreased quality of reconstructions, especially the accuracy of normals in the predicted surface. Finally, once both terms $\mathcal{L}_\text{img},\mathcal{L}_\text{norm}$ are removed, we observe that still, the single level-set consistency term is sufficient to boost the performance within the hybrid training. 
%

\begin{figure}[t]
\begin{center}
   \includegraphics[width=1.0\linewidth]{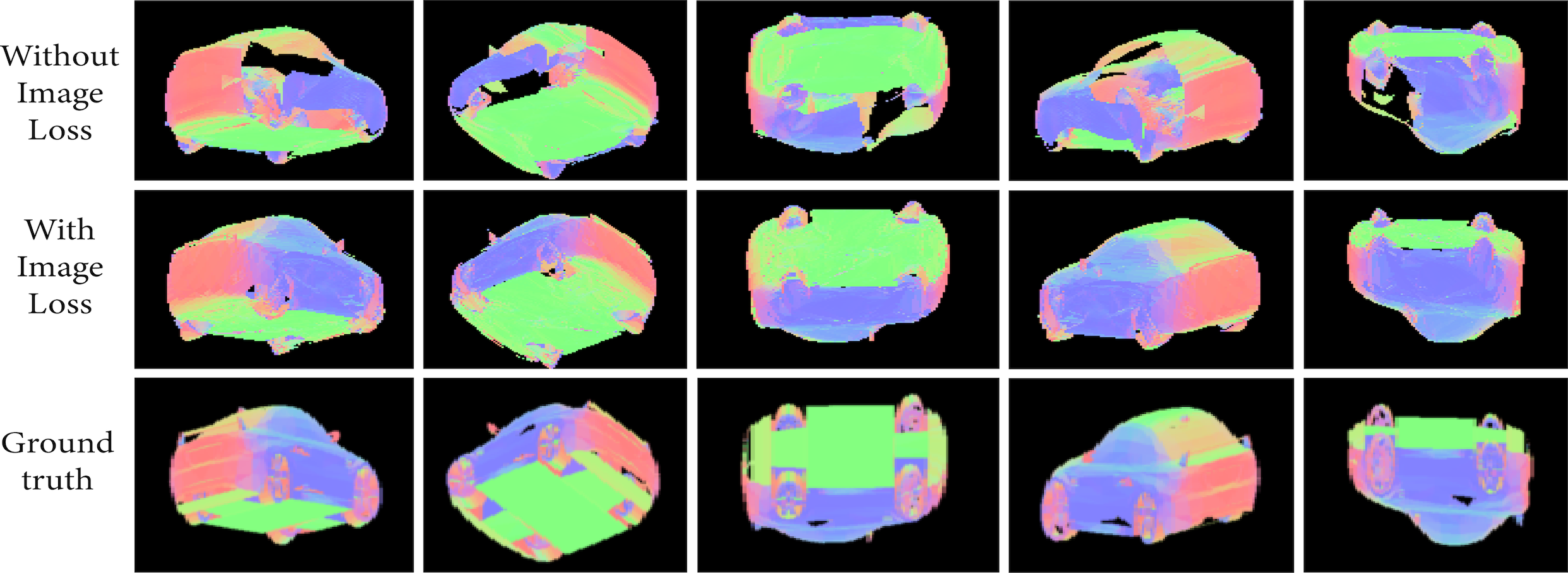}
\end{center}
\vspace{-0.2cm}
   \caption{Impact of the image loss. Rendered images from different viewpoints are shown for models trained with and without the image loss. Evidently, the image loss significantly improves the similarity of the output to the ground truth.}\label{fig:image}
\end{figure}

We also provide qualitative examples on how each loss term affects the quality of generated point clouds and meshes. Figure \ref{fig:image} illustrates impact of the image loss. Generated meshes from the AN branch are rendered from different viewpoints as shown in Figure \ref{fig:architecture}. Rendered images are colored based on per-face normals. As we observe, the image loss reduces artifacts such as holes, resulting in more accurate generation. {Note that our differentiable renderer uses backface culling, so the back side is not visible as the surface is oriented away from the camera.}

\begin{figure}[t]
\begin{center}
   \includegraphics[width=1.0\linewidth]{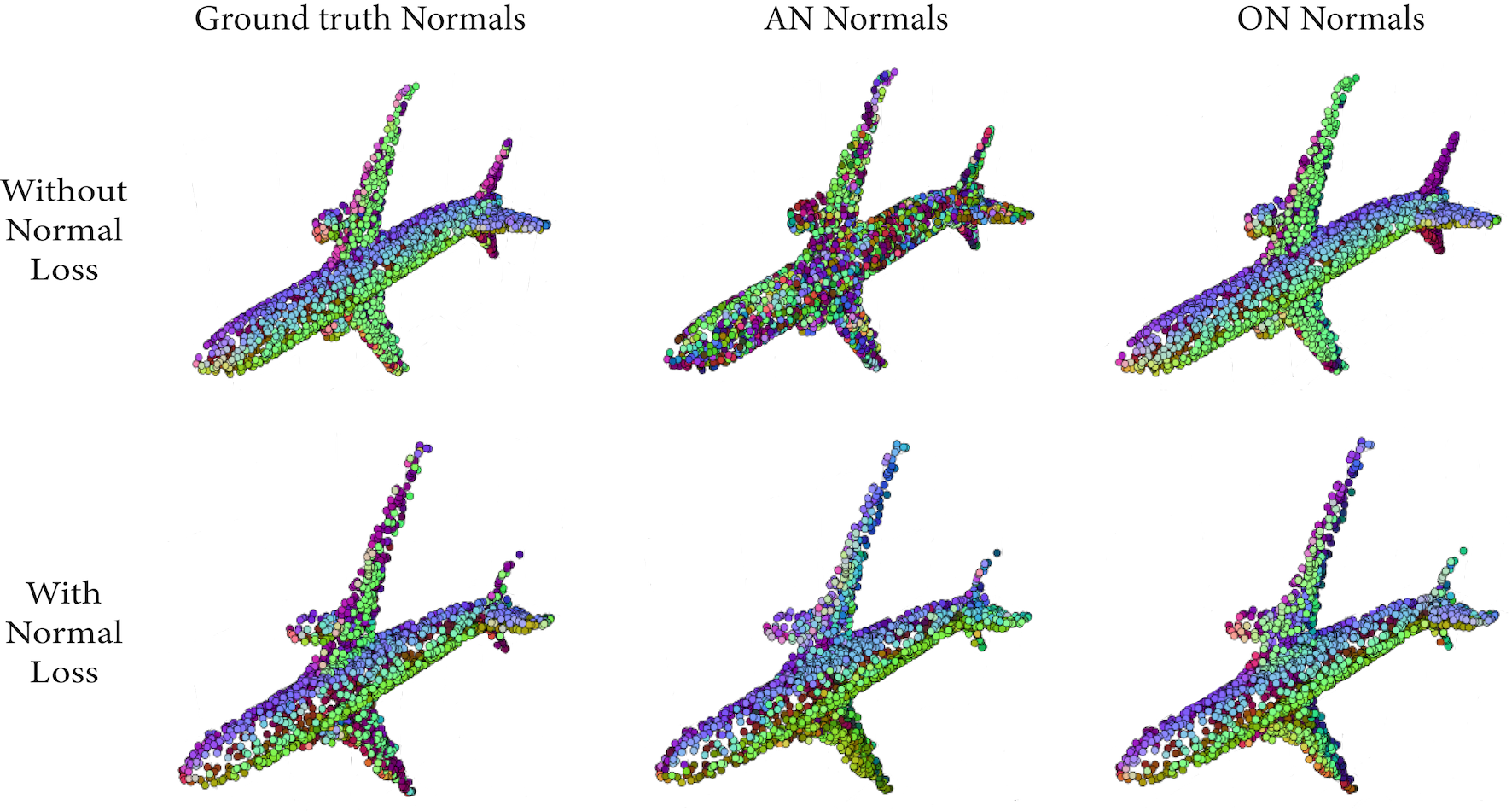}
\end{center}
\vspace{-0.2cm}
   \caption{Impact of the normal consistency loss. The generated point clouds are colored based on the ground truth normals, AtlasNet's (AN) normals, and OccupancyNet's (ON) normals. The results are then compared between a model trained with the normal consistency loss and a model trained without that loss; AN's normals significantly improve when the loss is incorporated, as it encourages alignment with ON's normals which tend to be close to the ground truth normals. }\label{fig:normal}
\end{figure}

\begin{figure}[!h]
\begin{center}
   \includegraphics[width=1.0\linewidth]{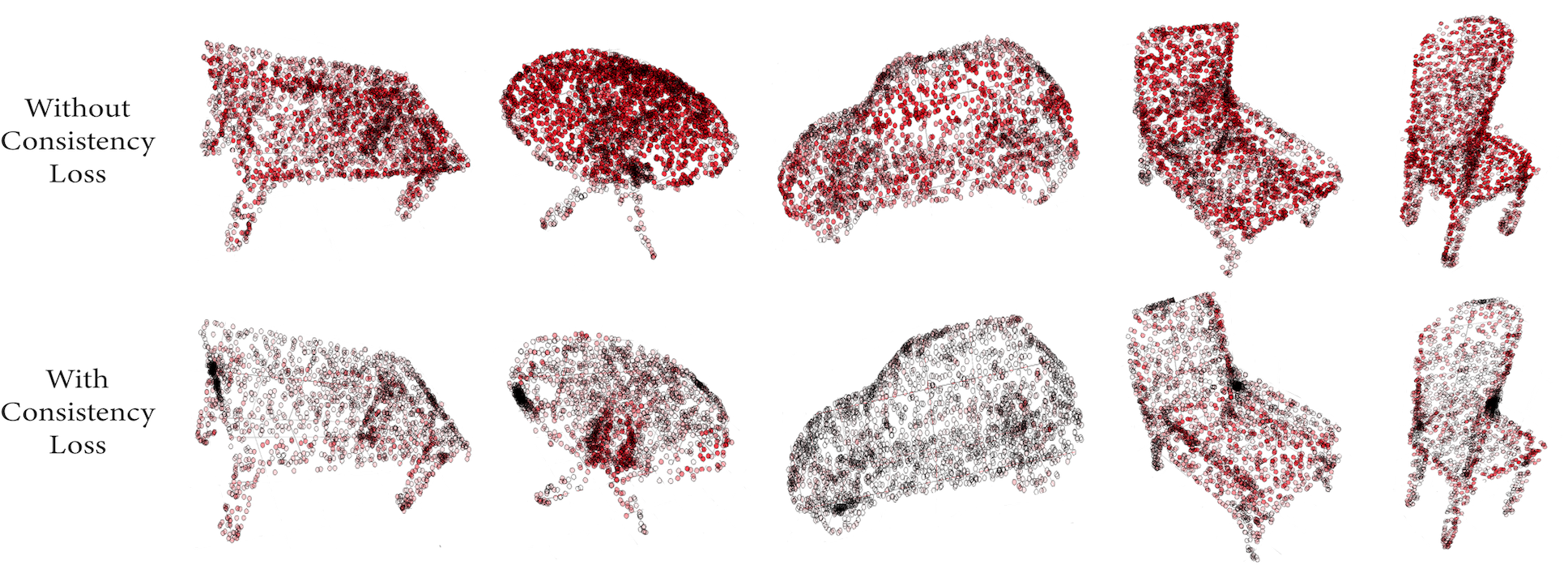}
   
   \vspace{0.1cm}
   
   \includegraphics[width=0.5\linewidth]{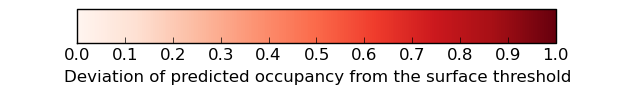} 
\end{center}
   \caption{Impact of the consistency loss. Each point in the generated point cloud is colored based on deviation of its predicted occupancy probability from the threshold $\tau$, with red indicating deviation.}\label{fig:cons}
\end{figure}

We next demonstrate the importance of the normal consistency loss in Figure \ref{fig:normal}. We colorize the generated point clouds (from the AN branch) with ground truth normals as well as normals computed from the AN and ON branches (Equation \ref{eq:atlas} and \ref{eq:occ}). We show the change in these  results between two models trained with and without normal consistency loss (Equation \ref{eq:normal}). As we observe, AN's normals are inaccurate for models without the normal loss. This is since the normals consistency loss drives AN's normals to align with ON's normals, which are generally close to the ground truth's.

Finally, Figure \ref{fig:cons} exhibits the effect of the consistency loss. We evaluate the resulting consistency by measuring the deviation from the constraint in Equation \ref{eq:constraint}, i.e., the deviation of the predicted occupancy probabilities from the threshold $\tau$, sampled on the predicted AtlasNet surface. We then color the point cloud such that the larger the deviation the redder the point is. Evidently, for models trained without the consistency loss this deviation is significantly larger, than when the consistency loss is incorporated.

\vpara{Surface reconstruction time. } One drawback of the implicit representations is the necessary additional step of extracting the level set. Current approaches require sampling a large number of points near the surface which can be computationally expensive. Our approach allows to circumvent this issue by using the reconstruction from the explicit branch (which is trained to be consistent with the level set of the implicit representation). We see that the surface reconstructing time is about order of magnitude faster for explicit representation (Table~\ref{tab:time}). Our qualitative and quantitative results suggest that the quality of the explicit representation improves significantly when trained with the consistency losses. 

\begin{table}[t]
\begin{center}
\begin{tabular}{|c|c|c|}
\hline
 & AN (explicit) & ON (implicit) \\
 \hline
Single-view Reconstruction & 0.037 & 0.400  \\
 \hline
 Auto-encoding & 0.025 & 0.428  \\
\hline
 \end{tabular}
\end{center}
\caption{Average surface reconstruction time (in seconds) for the explicit (AN) and implicit (ON) representations. Our approach enables to pick the appropriate reconstruction routine at inference time depending on the application needs, where the quality of the reconstructed surfaces increases due to dual training. } \label{tab:time}
\end{table}



\section{Conclusion and Future Work}
We presented a dual approach for generating consistent implicit/explicit surface representations using AtlasNet and OccupancyNet in a hybrid architecture via novel consistency losses that encourage this consistency. Various tests demonstrate that surfaces generated by our network are of higher quality, namely smoother and closer to the ground truth compared with vanilla AtlasNet and OccupancyNet.
A main shortcoming of our method is that it only penalizes inconsistency of the branches, but does not guarantee perfect consistency; nonetheless, the experiments conducted show that both representations significantly improve by using this hybrid approach during training.

We believe this is an important step in improving neural surface generation, and are motivated to continue improving this hybrid approach, by devising tailor-made encoders and decoders for both representations, to optimize their synergy. In terms of applications, we see many interesting future directions that leverage strengths of each approach, such as using AtlasNet to texture OccupancyNet's implicit level set, or using OccupancyNet to train AtlasNet's surface to encapsulate specific input points. 

\section{Appendix}
\begin{figure}[!h]
\begin{center}
\includegraphics[width=0.99\linewidth]{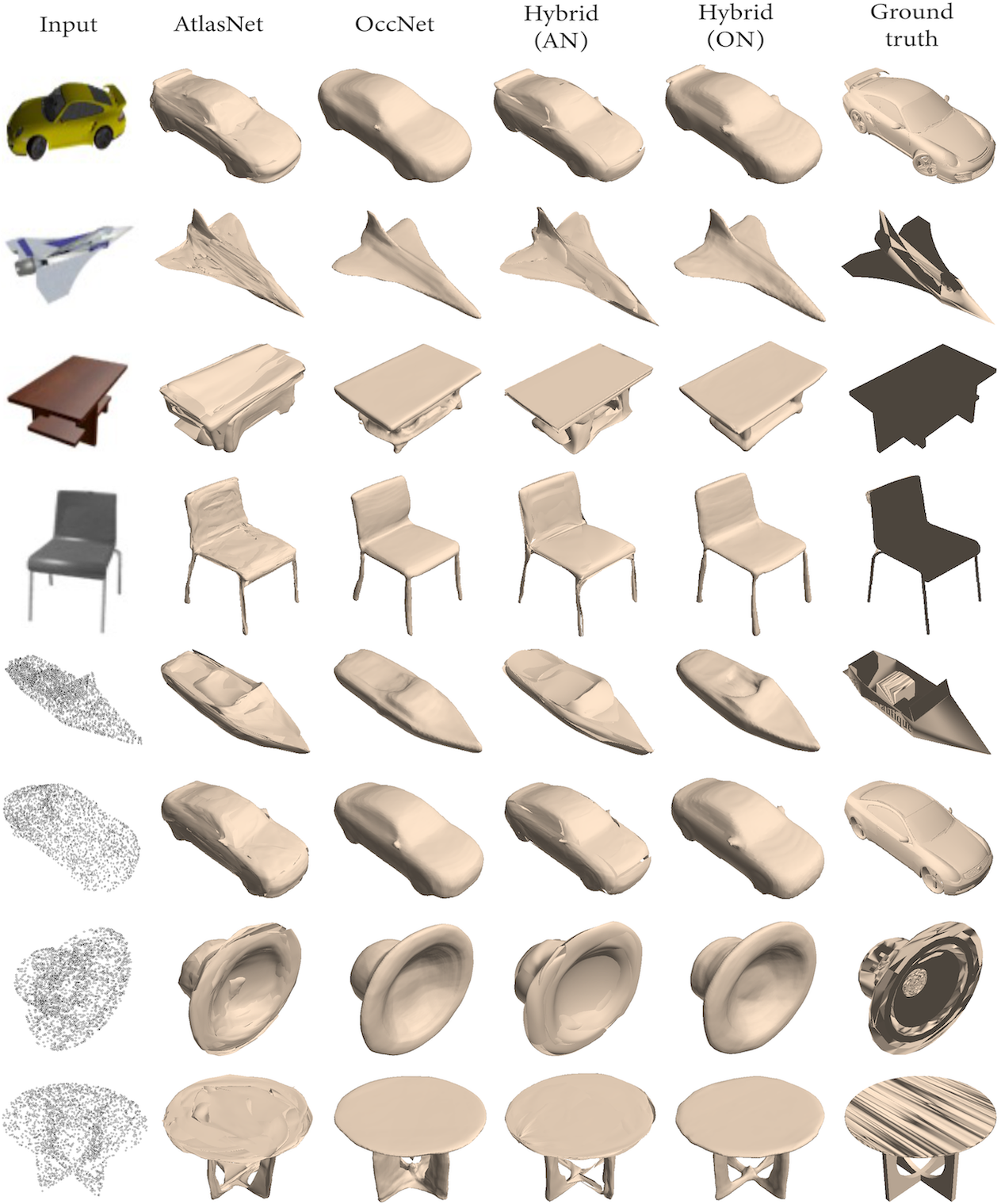}
\end{center}
   \caption{Random results on reconstructing surfaces from images and point clouds.}\label{fig:qual}
\end{figure}

 We visualize additional random results for single-view reconstruction and auto-encoding in Figure \ref{fig:qual}.

%
%
\bibliographystyle{splncs04}
\bibliography{egbib}
\end{document}